\documentclass{article}
\usepackage{amsmath,graphicx,mlspconf}
\usepackage{cite,url,amsthm, amssymb, mathtools,mathrsfs}
\usepackage{color}

\usepackage{mdwmath}
\usepackage{multirow,amsthm}
\usepackage{tikz}
\usetikzlibrary{plotmarks}
\usetikzlibrary{calc}
\usetikzlibrary{patterns}
\usetikzlibrary{shapes,arrows,snakes,backgrounds}

\usepackage{array,subfig}
\usepackage[english]{babel}

\title{Sparse Multi-Layer Image Approximation: Facial Image Compression}

\name{Sohrab Ferdowsi, Svyatoslav Voloshynovskiy, Dimche Kostadinov \thanks{The contacting author is Svyatoslav Voloshynovskiy, svolos@unige.ch}}

\address{University of Geneva\\
	Department of Computer Science\\
	Battelle B\^{a}t. A, 7 rte de Drize, 1227 Carouge, Switzerland\\ {http://sip.unige.ch/}  \\ \small{$\lbrace$Sohrab.Ferdowsi, svolos, Dimche.Kostadinov$\rbrace$@unige.ch }}
\begin{document}
%

\maketitle
\begin{abstract}
We propose a scheme for multi-layer representation of images. The problem is first treated from an information-theoretic viewpoint where we analyze the behavior of different sources of information under a multi-layer data compression framework and compare it with a single-stage (shallow) structure. We then consider the image data as the source of information and link the proposed representation scheme to the problem of multi-layer dictionary learning for visual data.  For the current work we focus on the problem of image compression for a special class of images where we report a considerable performance boost in terms of PSNR at high compression ratios in comparison with the JPEG2000 codec. 

\end{abstract}
\begin{keywords}
visual data representation, rate-distortion theory, lossy compression, image compression, dictionary learning
\end{keywords}
%
\section{Introduction}
\label{sec:intro}
Sparse data approximation is a fundamental problem in many areas of signal processing and machine learning. For different tasks like multimedia compression, content identification, multi-class classification and representation learning, one aims at straightforward, concise and computationally feasible approximations. 

The above requirements, while being conflicting in nature, have been formulated and extensively studied under different concepts and applications like rate-distortion theory, approximate nearest neighbor search, vector quantization, dictionary learning and supervised/unsupervised learning in different disciplines.

In this work, we try to address some of the issues considered in these topics by asking the question, which data representation scheme is the most concise in terms of memory storage, the fastest in terms of computational complexity and the most accurate in terms of fidelity to the original data.

To this end, we propose a framework that could potentially be used in many different applications such as quality enhancement, denoising, impainting, visual recognition and joint compression-encryption. The general idea behind this approach being present in several earlier works \cite{6781602}, \cite{Gersho92a}, we unify them together and treat the problem from a practically significant perspective along with an information-theoretic analysis. 

In particular, for this work, we consider the problem of image compression. We show that the proposed framework, when adapted to a particular class of images can gain a considerable compression performance increase compared to the JPEG2000 codec for the very low bit-rate regime on the images belonging to the same class, in our application, face images.

Such a problem formulation is of great practical significance for those applications where the significant amount of images of a similar nature, like facial/iris images in biometrics, medical images, remote sensing and astronomical images, are to be compressed and communicated. In this case, the usage of a generic codec whose basis vectors are not adapted to the statistics of image is known to be inefficient, especially in the low rate regime. In this case, the overhead for storing a common trained codebook might be minor in comparison to the gain for millions of images.


The paper is organized as follows. In section \ref{sec:shallowRD} we briefly review the classic Shannon Rate-Distortion theory where the data are represented in one single layer. In section \ref{sec:DeepIT} we discuss the information-theoretic analysis of the multi-layer structure. Section \ref{sec:Synthetic} studies the behavior of \textit{i.i.d.} sources of information under the multi-layer structure. Section \ref{sec:FaceImComp} considers images as the data to be treated within this framework where a short review of facial image compression in the literature is also provided. The experimental results for image compression are discussed in section \ref{sec:ExpRes}. Finally, we conclude the paper in section \ref{sec:Summary}.

\section{Shannon Rate-Distortion Theory: Shallow Representation}
\label{sec:shallowRD}

   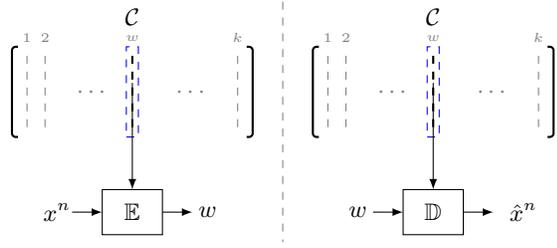
\begin{figure}  [!h]

\begin{center}
\begin{tikzpicture} [scale=0.4]

\node  at (-2.5-5,0) {\small{$x^n$}};
\draw[ -latex]	 (-2-5,0)  --  (-1-5,0);

\draw(-1-5,0.75) rectangle node{\small{$\mathbb{E}$}} (1-5,-0.75);
\draw[ -latex]	 (1-5,0) --(-3,0);
\node  at (-2.5,0) {\small{$w$}};

\node  at (0-5,6.5) {\small{$\mathcal{C}$}};
\draw[thick,rounded corners=0.5mm] (-3.8-5,5.5)--(-4-5,5.5)--(-4-5,2.5)--(-3.8-5,2.5);
\draw[thick,rounded corners=0.5mm] (3.8-5,5.5)--(4-5,5.5)--(4-5,2.5)--(3.8-5,2.5);
\draw[gray,dashed] (-3.5-5,5.2)--(-3.5-5,2.7);
\draw[gray,dashed] (-2.9-5,5.2)--(-2.9-5,2.7);
\node[gray] at (-1.3-5,4) {\small{$\cdots$}};
\draw[dashed,thick] (0-5,5.2)--(0-5,2.7);
\node[gray] at (2-5,4) {\small{$\cdots$}};
\draw[gray,dashed] (3.5-5,5.2)--(3.5-5,2.7);
\draw[dashed,blue] (-0.2-5,5.5) rectangle (0.2-5,2.5);
\draw[-latex] (0-5,4.3) -- (0-5,0.75);

\node[gray]  at (-3.5-5,5.8) {\tiny{$1$}};
\node[gray]  at (-2.9-5,5.8) {\tiny{$2$}};
\node[gray]  at (0-5,5.8) {\tiny{$w$}};
\node[gray]  at (3.5-5,5.8) {\tiny{$k$}};

\draw[gray,dashed] (0,7) -- (0,-1);

\node  at (2.5,0) {\small{$w$}};
\draw[ -latex]	 (3,0)  --  (-1+5,0);

\draw(-1+5,0.75) rectangle node{\small{$\mathbb{D}$}} (1+5,-0.75);
\draw[ -latex]	 (1+5,0) --(2+5,0);
\node  at (3+5,0) {\small{$\hat{x}^n$}};

\node  at (0+5,6.5) {\small{$\mathcal{C}$}};
\draw[thick,rounded corners=0.5mm] (-3.8+5,5.5)--(-4+5,5.5)--(-4+5,2.5)--(-3.8+5,2.5);
\draw[thick,rounded corners=0.5mm] (3.8+5,5.5)--(4+5,5.5)--(4+5,2.5)--(3.8+5,2.5);
\draw[gray,dashed] (-3.5+5,5.2)--(-3.5+5,2.7);
\draw[gray,dashed] (-2.9+5,5.2)--(-2.9+5,2.7);
\node[gray] at (-1.3+5,4) {\small{$\cdots$}};
\draw[dashed,thick] (0+5,5.2)--(0+5,2.7);
\node[gray] at (2+5,4) {\small{$\cdots$}};
\draw[gray,dashed] (3.5+5,5.2)--(3.5+5,2.7);
\draw[dashed,blue] (-0.2+5,5.5) rectangle (0.2+5,2.5);
\draw[-latex] (0+5,4.3) -- (0+5,0.75);

\node[gray]  at (-3.5+5,5.8) {\tiny{$1$}};
\node[gray]  at (-2.9+5,5.8) {\tiny{$2$}};
\node[gray]  at (0+5,5.8) {\tiny{$w$}};
\node[gray]  at (3.5+5,5.8) {\tiny{$k$}};

\end{tikzpicture}
 \end{center}
 \caption[example]  
 {Encoding and decoding in Shannon's shallow structure.}
  \label{fig:Shallow}
  \end{figure} 

The trade-off between the concise representation of a source of information and the fidelity is theoretically treated and formulated by Shannnon in \cite{shannon1959coding}. In this analysis, for the joint description of the outcomes of the sequence of random variables, $X^n = \lbrace X_1, \cdots X_n \rbrace$, the measure of compactness is the compression rate defined as $R_c = \frac{1}{n}  log_2{k}$ and measured in bits, if we store $k$ codewords in a codebook $\mathcal{C}$ that each of them refer to a data point in the space of $\mathcal{R}^n$.

The $k$ codewords, $\hat{x}^n$'s are generated from a distribution $p(\hat{x}^n|x^n)$ and organized into a shallow codebook $\mathcal{C}$ as shown in Fig. \ref{fig:Shallow}. Each codeword has the assigned index $1 \leq w \leq k$. This codebook $\mathcal{C} = \lbrace \hat{x}^n(1), \cdots,\hat{x}^n(k) \rbrace$ is shared between the encoder $\mathbb{E}$ and the decoder $\mathbb{D}$. The sequence $\hat{x}^n$ with an index $w$, represents a compressed counterpart of $x^n$. It should be pointed out that the codebook $\mathcal{C}$ is overcomplete, since $k = 2^{nR_c}$, i.e., the number of codewords $k$ is exponential in $n$. Moreover, the representation is sparse since only one codeword $\hat{x}^n(w)$ is used for the approximation of $x^n$. 

This representation leads to a loss of quality that should be measured as a distortion between $x^n$ and $\hat{x}^n$. One widely used measure of distortion between $x^n$ and $\hat{x}^n$ is the MSE, defined as $d(x^n,\hat{x}^n) = \frac{1}{n}\sum_{i=1}^n(x_i-\hat{x}_i)^2$.

The Shannon theory relates these two concepts by defining the rate-distortion function and relating it to the mutual information between the sequence and its representation; hence paving the way for calculation of this function for various sources. 

More concretely, the rate-distortion theory states that in order to guarantee to have the expected distortion between $x^n$ and $\hat{x}^n$, less than a threshold distortion value $D$, i.e., $ E[d(x^n,\hat{x}^n)] \leq nD$, the compression rate should be lower-bounded by the rate-distortion function $R_c(D)$. This lower bound is proven to be equal to:

\begin{equation}
R_c(D)=\min_{p(\hat{x}|x):\mathbb{E}[d(\hat{X},X)]\leq D} I(X;\hat{X}).
\label{eq:RD} 
\end{equation}

An important consequence of this theory states that, for memoryless sources of information emitting $\textit{i.i.d.}$ sequences, the distortion-rate function (an alternative to the rate-distortion function) is upper-bounded by that of the Gaussian source with the same variance $\sigma_x^2$, and MSE distortion measure, as:
\begin{equation}
D(R_c)\leq \sigma_x^2 \cdot 2^{-2R_c}.
\label{eq:DR_Gaussian} 
\end{equation}

These bounds, suggested by the Shannon's theory of rate-distortion, however, are proven to be achieved only for the asymptotic case where the block-length $n\rightarrow  \infty$. Consider for a fixed rate, any increase in the block-length $n$ would lead to an exponential increase in the number of representations as $k = 2^{nR_c}$. This means that, in the data representation language where several data points are to be stored in the memory and exhaustively matched in case queries are presented, one has to deal with an exponential complexity for both search and memory storage. Therefore, the current setup, while conceptually very important, cannot be appealing for many practical scenarios.
\section{Information-Theoretic Analysis of Multi-Layer Representation}
\label{sec:DeepIT}
Instead of the above single-layer (shallow) representation of information where we have a shallow codebook with $k \geq 2^{nI(\hat{X},X)}$ and $\hat{x}^n(w)$ is the $w^{\text{th}}$ representation vector in $\mathcal{C}$, consider the case where we have multiple codebooks $\mathcal{C}_1 \cdots \mathcal{C}_L$, where the $i^{\text{th}}$ codebook consists of $\mathcal{C}_i = \{ \hat {x}_i^n(1),\cdots,\hat {x}_i^n(k_i)\}$. The number of codewords, $k_i$, or the corresponding rates will be specified later.

Consider the final encoding or source approximation is done as $\hat{X}^n = \phi(\hat{X}^n_1,\cdots,\hat{X}^n_L)$. Therefore, the rate-distortion function can be calculated from equation (\ref{eq:RD}). The mutual information in this case can be bounded as 

\begin{align}
\begin{split}
I(X^n,\hat{X}^n) &= I(X^n,\phi(\hat{X}^n_1,\cdots,\hat{X}^n_L))\\ 
&\leq I(X^n,\hat{X}^n_1,\cdots,\hat{X}^n_L)) \\
&=\underbrace{I(X^n;\hat{X}^n_1)}_{nR_1}+\cdots+\underbrace{I(X^n;\hat{X}^n_L|\hat{X}_1^n + \cdots + \hat{X}_{L-1}^n)}_{nR_L} \\
&= \sum_{i=1}^L I(X^n;\hat{X}^n_i|\hat{X}_1^n + \cdots + \hat{X}_{i-1}^n) \\&= n(R_1 + \cdots+ R_L).
\end{split}
\label{eq:MI_Decomp} 
\end{align}

The important consequence of the developments in equation (\ref{eq:MI_Decomp}) is that, to achieve a high rate $R_c$ which requires exponential storage and computational complexities in the shallow representation (due to $k \geq 2^{nI(\hat{X},X)}$), one can achieve a targeted $R_c$ with $L$ codebooks each with very low rates such that
\begin{align*}
2^{nR_c} &= 2^{nR_1} \times \cdots \times 2^{nR_L} \\
&= 2^{n(R_1+\cdots R_L)},
\end{align*}
or equivalently,
\begin{equation}
R_c = \sum_{i=1}^L R_i.
\end{equation}
Therefore, the exponential nature of the required shallow codebook size for high rates is achieved by multiplication of smaller codebook sizes, i.e., the equivalent alphabet size will be:
\begin{equation}
k_ {eqq}= \prod_{i=1}^L k_i.
\end{equation}

   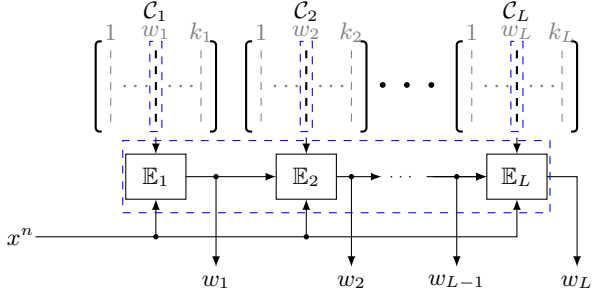
\begin{figure}  [!h]

\begin{center}
\begin{tikzpicture} [scale=0.4]




\node  at (-8.5,-2+1) { \small{$x^n$}  };
\draw[ -latex]	 (-8,-2+1)  --  (8,-2+1) -- (8,-0.75+1);  \filldraw  (-4,-2+1) circle (2pt);
\draw[ -latex]	 (-4,-2+1)  --   (-4,-0.75+1);    \filldraw  (1,-2+1) circle (2pt);
\draw[ -latex]	 (1,-2+1)  --   (1,-0.75+1);   

\draw(-5,0.75+1) rectangle node{\small{$\mathbb{E}_1$}} (-3,-0.75+1);
\draw[ -latex]	 (2,0+1)  -- (3.5,0+1); 
\draw[ -latex]	 (-2,1)  --  (-2,-2);  \filldraw  (-2,1) circle (2pt); \node  at (-2,-2.5) { \small{$w_1$}  };
\draw[ -latex]	 (-8+5,0+1)  --  (-5+5,0+1);
\draw(-5+5,0.75+1) rectangle node{\small{$\mathbb{E}_2$}} (-3+5,-0.75+1);
\draw[ -latex]	 (2.5,1)  --  (2.5,-2);  \filldraw  (2.5,1) circle (2pt); \node  at (2.5,-2.5) { \small{$w_2$}  }; 
\draw[ -latex]	 (5.2,0+1)  --  (7,0+1);
\draw[ -latex]	 (6,1)  --  (6,-2);  \filldraw  (6,1) circle (2pt); \node  at (6,-2.5) { \small{$w_{L-1}$}  }; 
\draw(-5+12,0.75+1) rectangle node{\small{$\mathbb{E}_L$}} (-3+12,-0.75+1);
\node  at (10,-2.5) { \small{$w_L$}  }; 
\draw[-latex] (9,0+1)--(10,0+1) -- (10,-2);
\node at (4.2,0+1) {\tiny{$\cdots$}};
\draw[ -latex]	 (5,0+1)  --  (7,0+1);

\draw[thick,rounded corners=0.5mm] (-5.8,5.5)--(-6,5.5)--(-6,2.5)--(-5.8,2.5);
\draw[thick,rounded corners=0.5mm] (-2.2,5.5)--(-2,5.5)--(-2,2.5)--(-2.2,2.5);
\draw[gray,dashed] (-5.5,5.2)--(-5.5,2.7);
\node[gray] at (-4.6,4) {\small{$\cdots$}};
\draw[dashed,thick] (-4,5.2)--(-4,2.7);
\node[gray] at (-3.1,4) {\small{$\cdots$}};
\draw[gray,dashed] (-2.5,5.2)--(-2.5,2.7);
 \draw[dashed,blue] (-4.2,5.5) rectangle (-3.8,2.5);
\draw[-latex] (-4,2.3)-- (-4,1.75);
\node[gray]  at (-5.5,5.8) {\small{$1$}};
\node[gray]  at (-4,5.8) {\small{$w_1 $}};
\node[gray]  at (-2.5,5.8) {\small{$k_1 $}};

\draw[thick,rounded corners=0.5mm] (-5.8+5,5.5)--(-6+5,5.5)--(-6+5,2.5)--(-5.8+5,2.5);
\draw[thick,rounded corners=0.5mm] (-2.2+5,5.5)--(-2+5,5.5)--(-2+5,2.5)--(-2.2+5,2.5);
\draw[gray,dashed] (-5.5+5,5.2)--(-5.5+5,2.7);
\node[gray] at (-4.6+5,4) {\small{$\cdots$}};
\draw[dashed,thick] (-4+5,5.2)--(-4+5,2.7);
\node[gray] at (-3.1+5,4) {\small{$\cdots$}};
\draw[gray,dashed] (-2.5+5,5.2)--(-2.5+5,2.7);
 \draw[dashed,blue] (-4.2+5,5.5) rectangle (-3.8+5,2.5);
\draw[-latex] (-4+5,2.3)--(-4+5,0.75+1);
\node[gray]  at (-5.5+5,5.8) {\small{$1$}};
\node[gray]  at (-4+5,5.8) {\small{$w_2 $}};
\node[gray]  at (-2.5+5,5.8) {\small{$k_2 $}};
\node at (4.5,4) {\huge{$\cdots$}};

\draw[thick,rounded corners=0.5mm] (-5.8+12,5.5)--(-6+12,5.5)--(-6+12,2.5)--(-5.8+12,2.5);
\draw[thick,rounded corners=0.5mm] (-2.2+12,5.5)--(-2+12,5.5)--(-2+12,2.5)--(-2.2+12,2.5);
\draw[gray,dashed] (-5.5+12,5.2)--(-5.5+12,2.7);
\node[gray] at (-4.6+12,4) {\small{$\cdots$}};
\draw[dashed,thick] (-4+12,5.2)--(-4+12,2.7);
\node[gray] at (-3.1+12,4) {\small{$\cdots$}};
\draw[gray,dashed] (-2.5+12,5.2)--(-2.5+12,2.7);
 \draw[dashed,blue] (-4.2+12,5.5) rectangle (-3.8+12,2.5);
\draw[-latex] (-4+12,2.3)--(-4+12,0.75+1);
\node[gray]  at (-5.5+12,5.8) {\small{$1$}};
\node[gray]  at (-4+12,5.8) {\small{$w_L $}};
\node[gray]  at (-2.5+12,5.8) {\small{$k_L $}};


\node  at (-4,6.5) {\small{$\mathcal{C}_1$}};
\node at (-4+5,6.5) {\small{$\mathcal{C}_2$}};
\node  at (-4+12,6.5) {\small{$\mathcal{C}_L$}};

\draw[dashed,blue] (-5.1,1.2+1) rectangle (9.1,-1.2+1);
\end{tikzpicture}
\end{center}
 
\caption[example] {The encoding structure in the multi-layer representation. $\mathbb{E}_i$, in the general formulation, has the knowledge of $\hat{x}_1^n, \cdots \hat{x}_{(i-1)}^n$.}
\label{fig:MLayers} 
  \end{figure} 

Fig. \ref{fig:MLayers} sketches the structure of codebooks and decoding in the multi-layer representaion.

\subsection{Multi-Layer Additive Structure}
\label{subsec:Additive}



Suppose the special case when we have the reconstruction function $\phi(\cdot)$ to be additive, i.e., $\hat{X} = \hat{X}_1 + \cdots + \hat{X}_L$. Given a realization of the source, the decoder in this case consists of finding the Euclidean nearest neighbor of the sequence within the codewords of the first codebook, calculating the error of estimation and passing it to the next stage and repeating the same procedure until the last stage where the overall error will be equivalent to the error in the last stage since, $\hat{X}_1^n = \mathcal{Q}_1(X^n), \hat{X}_2^n = \mathcal{Q}_1(X^n-\hat{X}_1^n), \cdots , \hat{X}_L^n = \mathcal{Q}_L(X^n-(\hat{X}_1^n + \cdots + \hat{X}_{L-1}^n))$, where $\mathcal{Q}_i(\cdot)$ denotes the Vector Qunatizer for the $i^\text{th}$ stage with distortion $D_i$. 

Moreover, for the Gaussian source $X_i \sim \mathcal{N}(0,\sigma_x^2)$ we have that $R_1 = \frac{1}{2}log_2^+(\frac{\sigma_x^2}{D_1})$, $R_2 = \frac{1}{2}log_2^+(\frac{D_1}{D_2}), \cdots R_L = \frac{1}{2}log_2^+(\frac{D_{L-1}}{D_L})$ where $D_i$ stands for the corresponding distortion, and $D_L = D$.

The decoding as well as memory complexities will be $O(\sum \limits_{i=1}^k 2^{nR_i})$ instead of $O(2^{nR_c})$ in the shallow structure. 

Reducing the multi-stage encoding function $\phi(\cdot)$ to the addition operator, while being simple and intuitive, reduces the optimality since in general
\begin{equation*}
h(\hat{X}_1 + \cdots + \hat{X}_L) \leq h(X_1 , \cdots , X_L)
\end{equation*}
and $I(X;\hat{X}_1 + \cdots + \hat{X}_L)$ cannot be decomposed directly to conditional terms as in equation (\ref{eq:MI_Decomp}).

This issue, while reducing optimality due to some information loss in the addition operations, keeps the great advantage of breaking the exponential complexities of one huge shallow structure to several codebooks of considerably smaller sizes. In addition, a practical question of learnability of exponential codebook in the shallow structure is infeasible and requires also exponential number of training samples. In contrast, the multi-layer structure can be easily trained for low rates $R_i$.

In section \ref{sec:Synthetic}, we simulate the performance of this scheme for \textit{i.i.d.} sources of information and show that this loss is not a limiting factor.
\section{Multi-Layer Representation of $\textit{i.i.d.}$ Sources for Synthetic Data}
\label{sec:Synthetic}
Consider the stationary ergodic source $X^n$ with $X_i \sim p_X(x)$. The realizations of this source are to be represented with codebooks $\mathcal{C}_1,\cdots,\mathcal{C}_L$ each with $k_i$ codewords. The decoding is done as described in section \ref{subsec:Additive}. 

For the encoding at each stage, using the stationarity and ergodicity assumptions on the source and therefore the specific geometry imposed to the data distribution in $\mathcal{R}^n$ as $n$ grows large, we design the codewords in different codebooks very efficiently using only \textit{random codewords} that are properly normalized.

Suppose the case where $X_i \sim \mathcal{N}(0,\sigma_x^2)$ for all $1 \leq i \leq n$, which means that the data are $\textit{i.i.d.}$. In this case, the data is concentrating around a spherical shell with radius $\rho = \sqrt{n\sigma_x^2}$, as $n$ grows large. 

For the first stage of encoding, suppose we want to compress the data to the rate $R_1$. The achievable distortion for this stage and $n$ large enough is given by equation (\ref{eq:DR_Gaussian}) as $D_1=\sigma_x^2 2^{-2R_1}$ which is achieved for optimal codebook design, in this case with random structure. 

Due to the optimality proved for this hypothetical case in terms of MSE distortion, one can conclude the orthogonality of the vector of estimation with its error (due to the principle of orthogonality), i.e.,
\begin{equation*}
\big\langle (x^n-\hat{x}_1^n),\hat{x}_1^n \big\rangle \simeq 0.
\end{equation*}
Therefore, from the law of cosines, one can confirm that the variance of the codewords of the first codebook is:
\begin{equation}
\sigma_{\hat{x},1}^2 = \sigma_x^2 \big( 1-2^{-2R_1} \big).
\label{eq:Variances1}
\end{equation}
Extending the argument to other layers, it can be concluded that the variance of the codewords of the $i^{\text{th}}$ layer is given as:
\begin{equation}
\sigma_{\hat{x},i}^2 = \sigma_{\hat{x},i-1}^2\big(1-2^{-2R_i}\big).
\label{eq:Variances2}
\end{equation}
\subsection{Design of Random Codebooks for Multi-Layer Representation}
The use of random codebooks is appealing both in theory and for practical applications. Avoiding overfitting to the seen data in a machine learning setup, preservation of privacy and security in multimedia or medical data management and eliminating the computational cost of codebook design are among the advantages of this approach. 
 
\begin{figure}  [!h]
   \begin{center}
   \begin{tabular}{c}
    \subfloat[distortion vs. variance]{\includegraphics[width=0.25\textwidth]{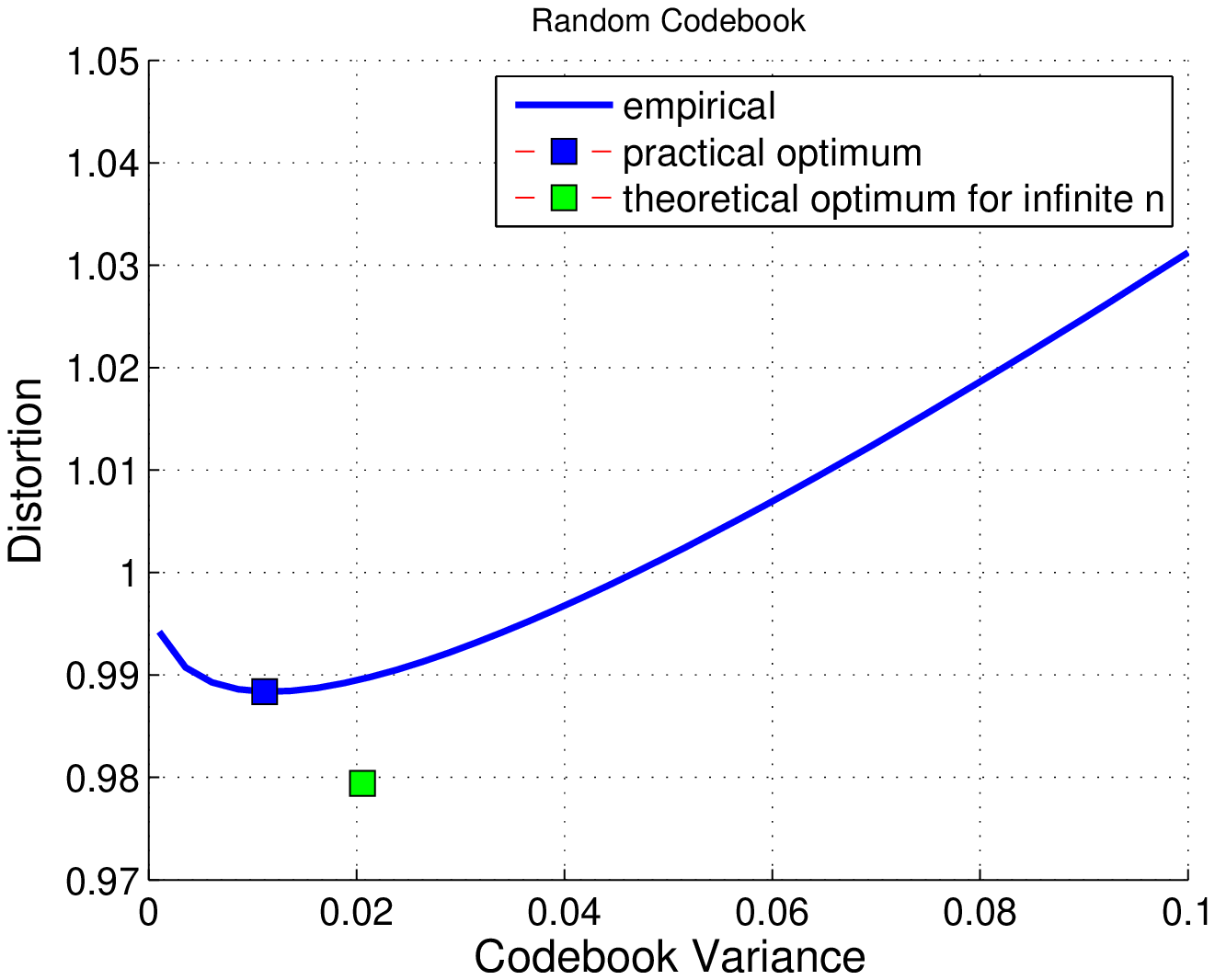}\label{subfig:D-vs-CVariance}}
    \subfloat[orthogonality vs. variance]{\includegraphics[width=0.25\textwidth]{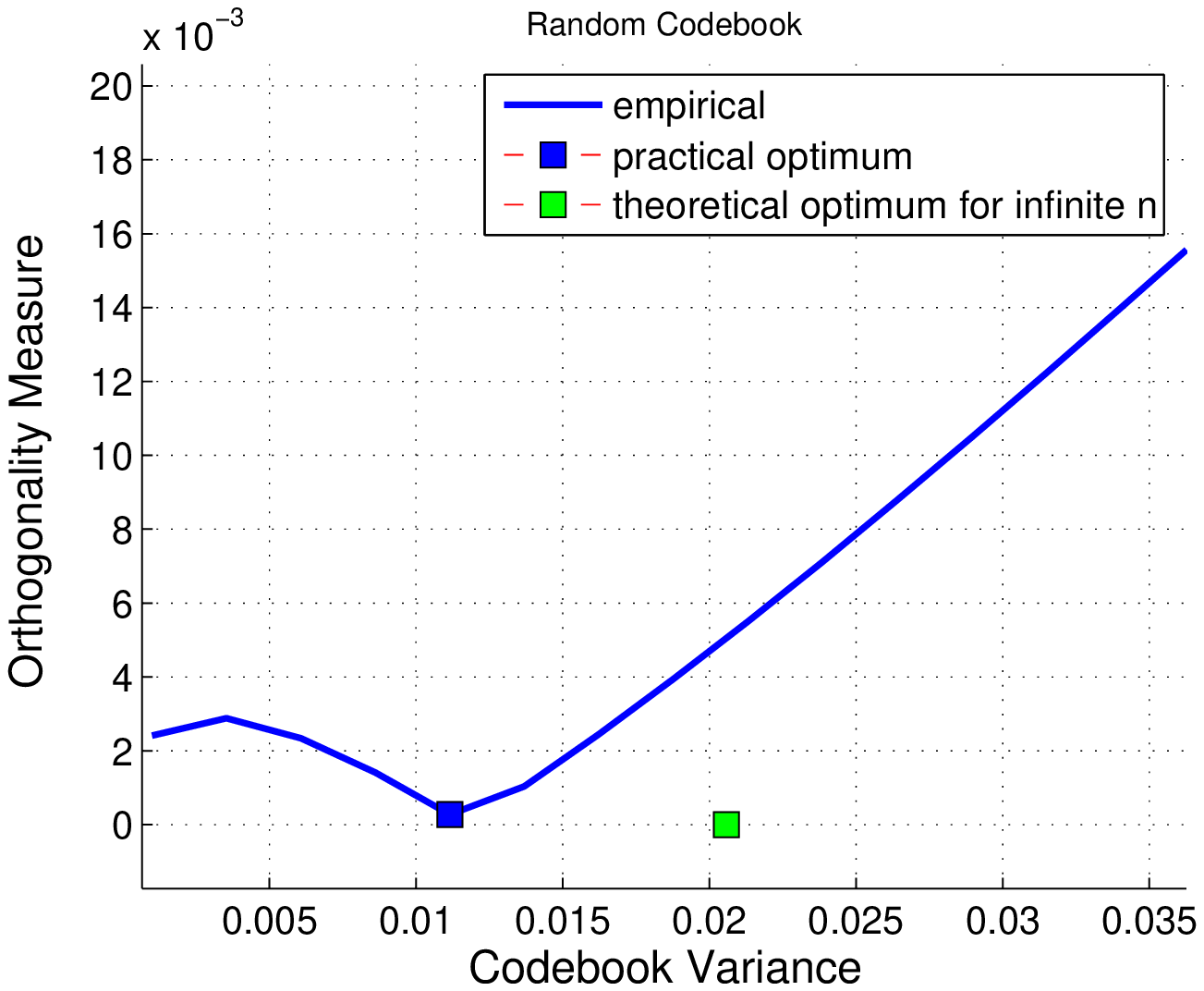}\label{subfig:OrthTest}}
   \end{tabular}
   \end{center}
   \vspace{-0.5cm}
\caption[example]{The effect of codebook normalization on the achieved distortion and orthogonality for a zero-mean Gaussian source in $\mathcal{R}^{n}$with $n=200$ and $\sigma_x^2 = 1$ and a randomly generated codebook with $k=8$ with varying variance.}
{ \label{fig:normalization}}
   \end{figure}

To this end, equations (\ref{eq:Variances1}) $\&$ (\ref{eq:Variances2}) should be considered in random codebook design. Fig. \ref{fig:normalization} shows the effect of normalization of codewords in a codebook on the achieved distortion and also orthogonality as measured by $E[\big\langle(X^n-\hat{X}^n),X^n \big\rangle]$ for different codebook variances.

As is shown in the figure, the empirical optimum for $n=200$ is not far from the theoretical optimum when $n \rightarrow \infty$, the difference being due to geometrical variations of the n-sphere for different values of $n$. 
 
 In fact, by proper normalization, as dictated by these equations, we show that we can get very close to the theoretical distortion-rate limit of equation (\ref{eq:DR_Gaussian}) for moderate values of $n$.

\begin{figure}  [!h]
   \begin{center}
   \begin{tabular}{c}
   \includegraphics[width=0.45\textwidth]{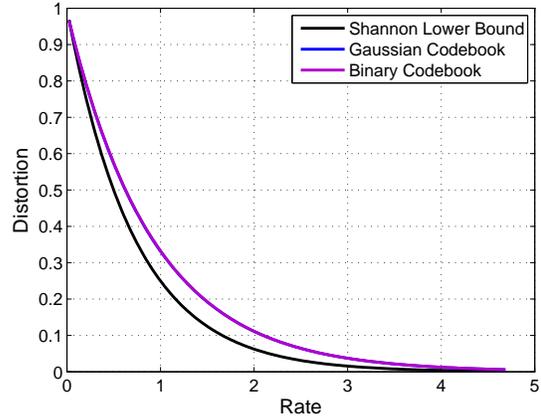}
   \end{tabular}
   \end{center}
   \caption[example] 
   { \label{fig:R-D} 
Distortion-rate behavior of multi-stage additive structure for an \textit{i.i.d.} Gaussian source with Gaussian and binary codebooks. The data dimension was $n=512$ and $k_i=2^{12}$ for all values of $1 \leq i \leq L=200$. Therefore, the rate at each stage was $R_i \simeq 0.023$ with an overall rate of $R_c = 4.69$.}
   \end{figure}

Fig. \ref{fig:R-D} shows the achieved distortion for \textit{i.i.d.} source $X^n$ with $n=512$ synthesized from $X \sim \mathcal{N}(0,1)$ for different compression rates. We consider the compression using two different sets of codebooks. First is a randomly generated \textit{i.i.d.} Gaussian with $\hat{X}_i \sim \mathcal{N}(0,D_{i-1}-D_i)$ for the $i^{\text{th}}$ layer with $D_0 = \sigma_x^2 $ where $D_{i-1}$ is the average distortion of the $(i-1)^{\text{th}}$ layer.

The second set of codebooks are equiprobable binary codebooks with alphabet $\hat{\mathcal{X}}_i = \lbrace \pm \alpha_i \rbrace$, $\alpha_i$ chosen to guarantee the same variance as the Gaussian case in each layer.

As is seen from the figure, the achieved distortion-rate function, without the exponential complexity burdens of the shallow structure, closely approximates the behavior of the Shannon lower bound in equation (\ref{eq:DR_Gaussian}). The difference with the theoretical limit is due both to the finite block length and the information loss due to the additive encoding, as explained in section \ref{subsec:Additive}. 

Interestingly, the behavior of the two codebook design strategies is the same. The reason is due to the choice of very small rates at each stage. In fact, in an analogy with channel coding, the dual problem of rate-distortion theory, one can verify that capacities of the Gaussian channel and binary symmetric channel are very close at extremely small rates.

This fact is of very much practical significance, since, if the rate selection and normalization is done properly, one does not have to worry about matching the distribution of the codebooks with that of the source. Moreover, the memory storage of real valued codewords can be reduced to that of binary values.  
\section{Facial Image Compression}
\label{sec:FaceImComp}
Due to its  practical significance, image compression has become a very mature field of both research and technology. Among the existing methods of image compression, JPEG2000 is reported to be among the best existing algorithms used in practice \cite{TaubmanMarcellin201306} with a very intricate structure to achieve a highly optimized trade-off between compression ratio and performance  . However, since it is a general purpose codec, for applications where compression of a large amount of similar images is concerned, one could think of methods of compression to use the extra redundancy present due to the similarities of application images. Moreover, the JPEG2000 codec is not capable of providing very high compression ratios while many applications would require images to be highly compressed and compromise quality for description efficiency.

One important example for this scenario is the compression of facial images. They are available in large quantities in big databases of police departments, organizations and entities with lots of employees and users. Efficient compression of these images in terms of storage and computational complexity is very important since it will result immediately in more resources and thus providing services to more users. Moreover, in some applications, rather than quality and fine details, the recognition informativeness of facial images is of more importance.

Apart from the very numerous literature in image compression, there has been several works on compression of facial images. In \cite{4286990}, a facial compression scheme based on Vector Quantization was proposed where a considerable performance improvement over the JPEG2000 was reported at very low rates. However, this method needs detection of facial features (sometimes manually) and alignment by geometrical transformation into a canonical form and also background removal which makes it very sensitive to the required pre-processing. Within the same setup, an approach based on dictionary learning with the K-SVD algorithm was proposed in \cite{Bryt2008270} where a special dictionary was learned for every block location of the image. In another work \cite{6844846}, a facial image compression using Redundant Tree-Based Wavelet Transform (RTBWT) was used with the same pre-processing and a filtering-based post-processing to improve the quality of images. In spite of their high performance in terms of PNSR, the problem with these approaches is that they rely very much on the alignment of images and they are less likely to generalize once the imaging setup is changed a bit.

Another scheme was proposed in \cite{5740941} where the authors propose a codec by using the Iteration Tuned and Aligned Dictionary (ITAD) to compress facial images where dictionaries are tuned in every iteration of the pursuit algorithm used. A considerable compression performance gain is reported for a wider range of compression rates. However, the tree structure of the dictionaries will require a considerable storage. 

\subsection{Multi-layer approximation of Images} 
We apply the above framework to compress facial images. Images from the training set are divided into non-overlapping blocks and then gathered in a database. Without any special pre-processing, the blocks are vectorized and fed to the simple k-means algorithm. The residual of quantization is fed to the next stages for further quantization. To avoid over-fitting, $k_i$, the number of cluster centroids (codewords) at the $i^\text{th}$ layer is chosen such that the distortion of reconstruction of the test data is within a margin from the distortion of the reconstruction of the training data. 

The encoding part consists of assigning to each image block a sequence of indices $\lbrace w_i \rbrace_{i=1}^L$ each taking values from an alphabet of $k_i$ codewords. Therefore, the Bits Per Pixel (BPP) value for the image will be $ \sum_{i=1}^L log_2(k_i)/(b^2)$ where $b$ is the block size. This value could be reduced by the use of an entropy coding applied to indices where a probability table could be trained from the training set for each of the stages.

The decoding part simply consists of table look-ups to read the values of the corresponding entropy-decoded sequences of codewords for each block and their addition. This process could be done online and sequentially once the required bits for each stage is received. 
\section{Experimental Results for Image Compression}
\label{sec:ExpRes}

We used 2400 randomly chosen images ($80 \%$ for training and $20\%$ for testing) from the \textit{CroppedYaleB}\cite{GeBeKr01} database of cropped facial images with different lighting conditions. This is a difficult database for compression since the variation of lighting in images is very significant and shadows could obscure different parts of face in different images. Therefore, one cannot train highly specialized dictionaries for different locations. Moreover, unlike the databases used in the existing approaches, the background is completely removed from the faces and the algorithm cannot favor from the redundant areas common in all images.  

\begin{figure}  [!h]
   \begin{center}
   \begin{tabular}{c}
    \subfloat[36 of 256 codewords- $1^{\textbf{st}}$ layer]{\includegraphics[width=0.25\textwidth]{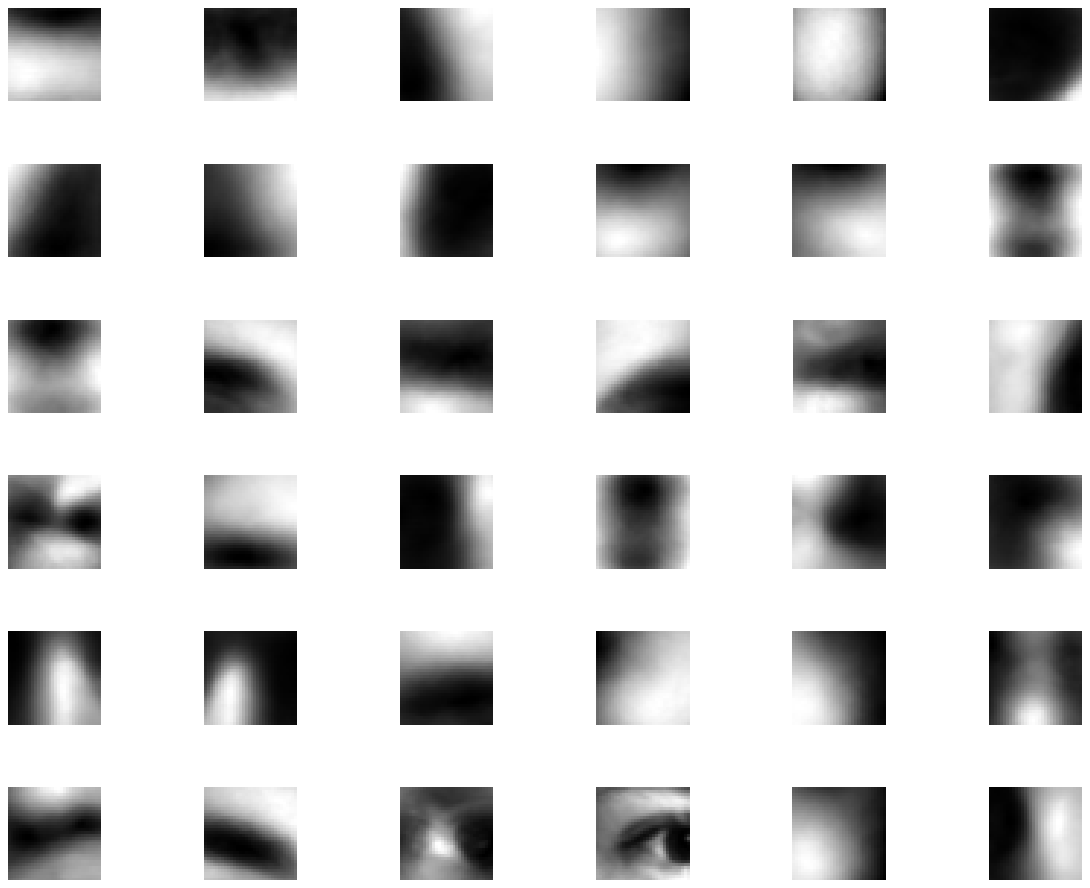}}
    \subfloat[16 codewords- $19^{\textbf{th}}$ layer]{\includegraphics[width=0.25\textwidth]{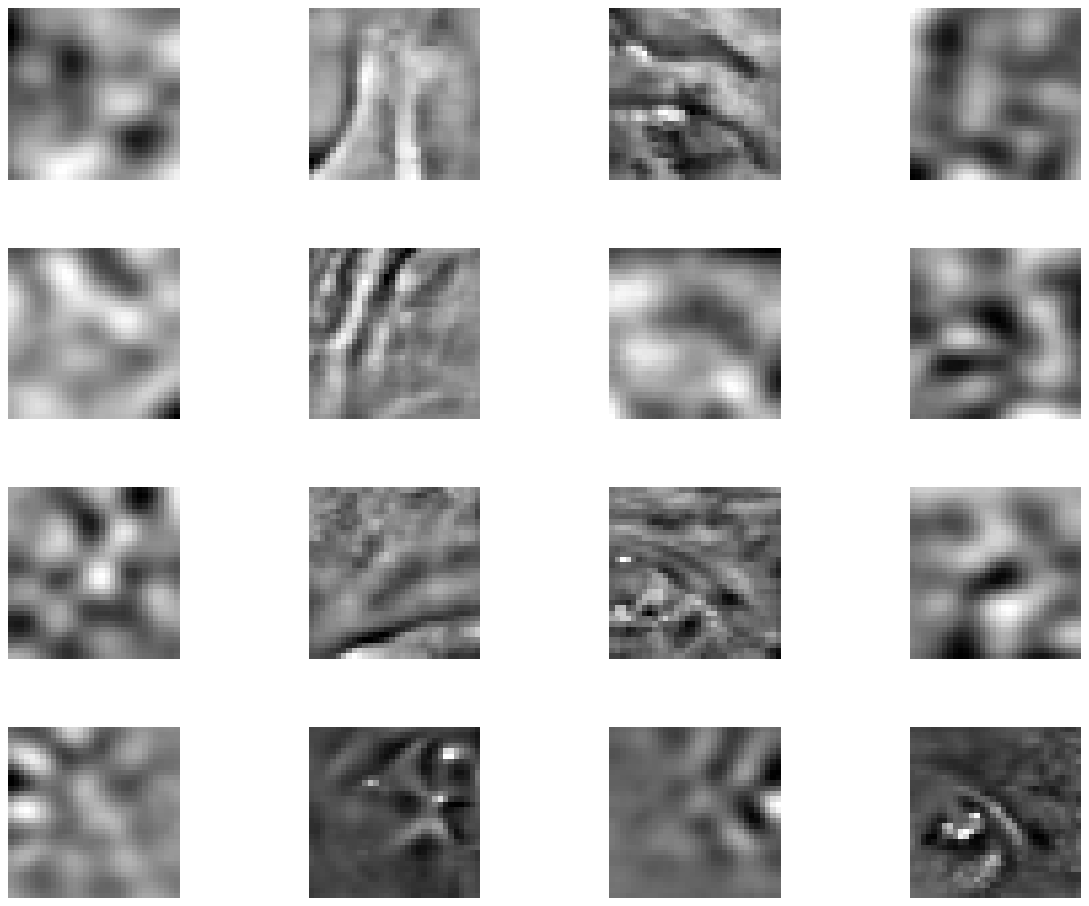}}
   \end{tabular}
   \end{center}
   \caption[example] {Randomly selected codewords from two different layers.}
   { \label{fig:codewords}  }
   \end{figure}

We used $L=20$ layers of global codebooks with $k=256, 128, 32$ and $16$ codewords at the first, second, third and forth consecutive five layers, respectively. As was previously mentioned, the choice of these values should avoid over-fitting. As is understandable from the values chosen for $k_i$, also as seen from Fig. \ref{fig:codewords}, the latter stages have less correlated structure and tend to over-train more easily.

\begin{figure}  [!h]
   \begin{center}
   \begin{tabular}{c}
\hspace{-1cm}       
       \subfloat[]{\includegraphics[width=0.25\textwidth]{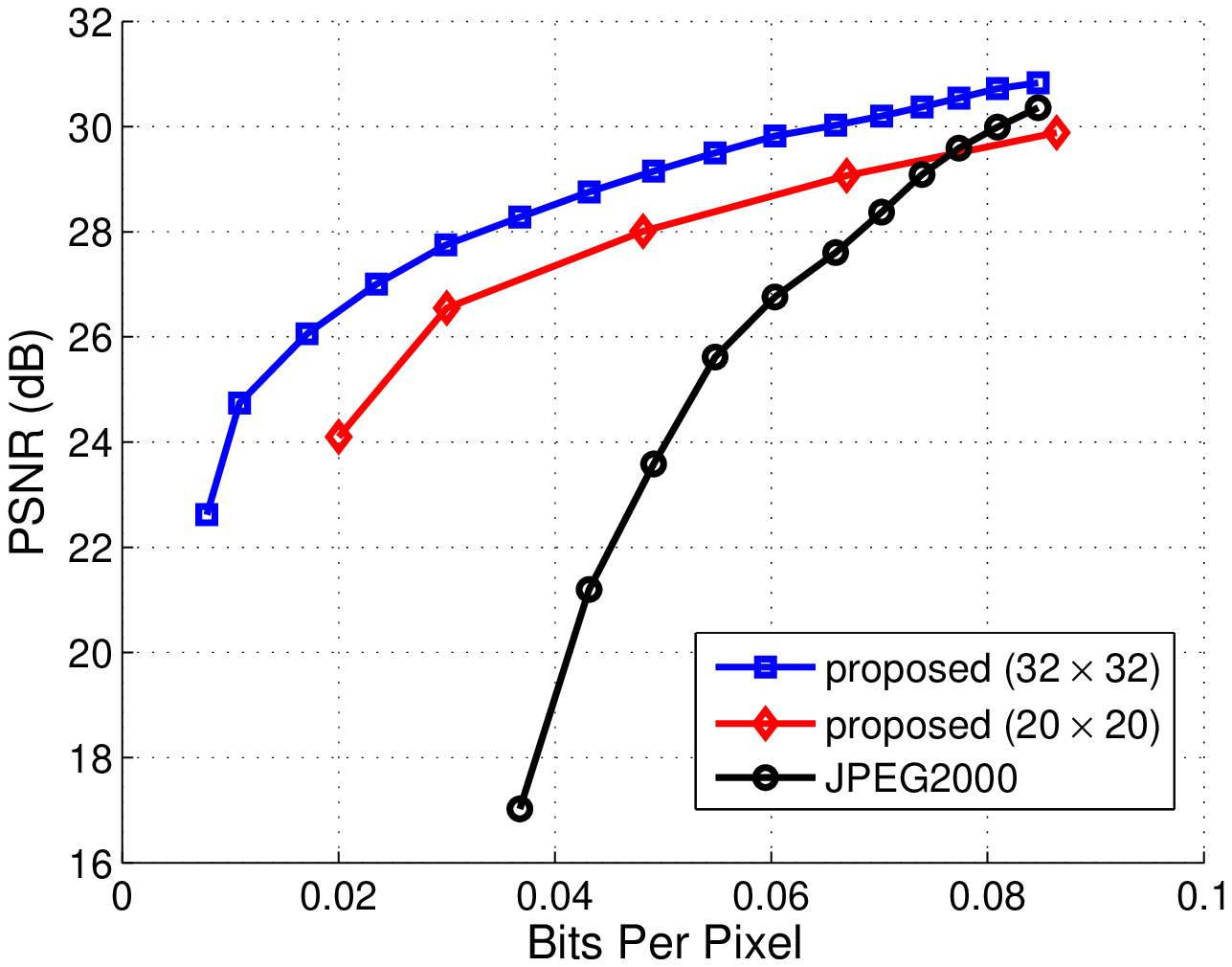}\label{subfig:PSNR}}
    \subfloat[]{\includegraphics[width=0.25\textwidth]{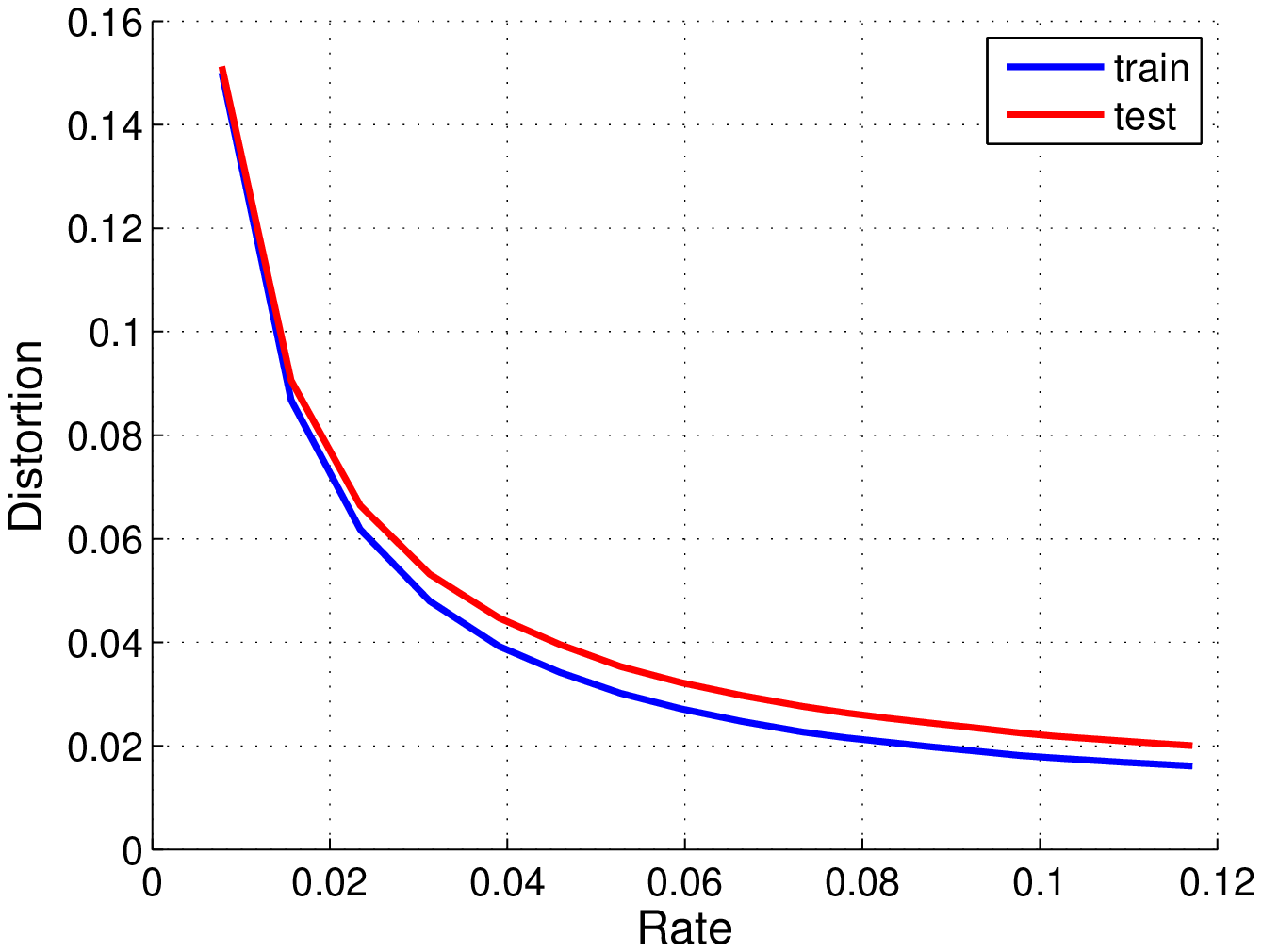}\label{subfig:R-D images}}
   \end{tabular}
   \end{center}
   \caption[example] {average compression performance over 480 test facial images of the \textit{CroppedYale} set (a) comparison with JPEG2000 in terms of PSNR (b) distortion-rate behavior (train set size was 1920).}}
   { \label{fig:PSNR} 
   \end{figure}

Fig. \ref{subfig:PSNR} sketches the compression behavior of this experiment with that of the JPEG2000 codec in terms of PSNR(dB) for different values of Bits Per Pixel (BPP) and Fig. \ref{subfig:R-D images} shows the rate-distortion performance. As is seen from this figure, the quality of the compressed facial images at very low rates is significantly superior to that of the JPEG2000. We used a simple arithmetic coding for the indices.

\begin{figure}  [h!]

\begin{center}
  \begin{tabular}{lcl}

\hspace{-1.3cm}  
    \subfloat[original]{\includegraphics[width=0.2\textwidth]{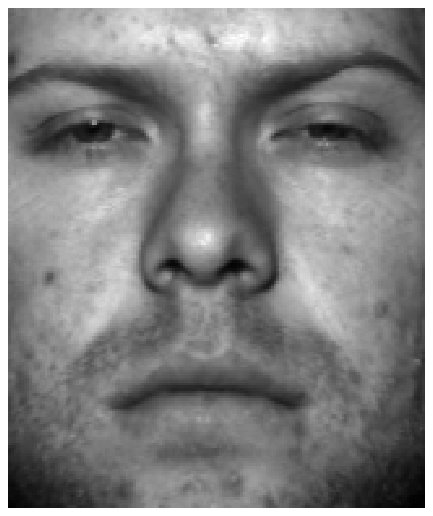}} &

\hspace{-1.0cm}    
    \subfloat[proposed (bpp $=0.05$)]{\includegraphics[width=0.2\textwidth]{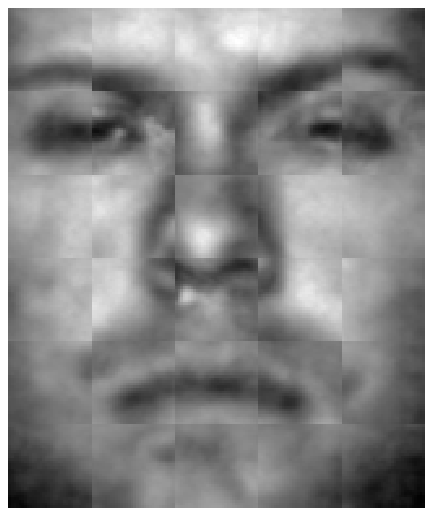}} &
    
\hspace{-.7cm}   
     \subfloat[JPEG2000 (bpp $=0.05$)]{\includegraphics[width=0.2\textwidth]{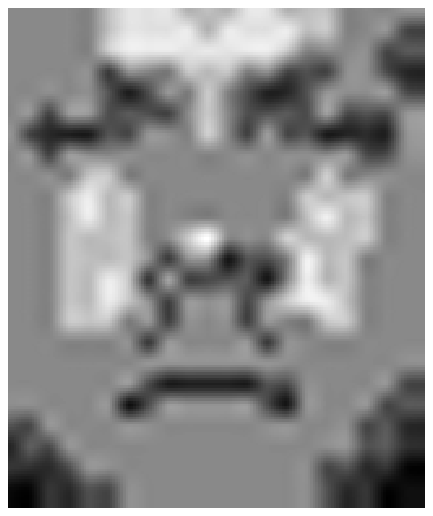}} 
\\  
 
\hspace{-1.3cm}    
    \subfloat[original]{\includegraphics[width=0.2\textwidth]{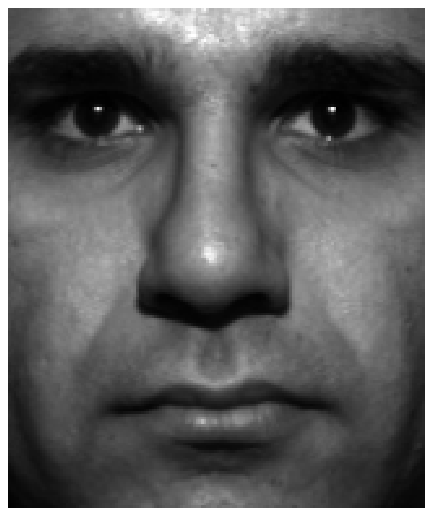}} &
  
\hspace{-1.0cm}    
    \subfloat[proposed (bpp $=0.09$)]{\includegraphics[width=0.2\textwidth]{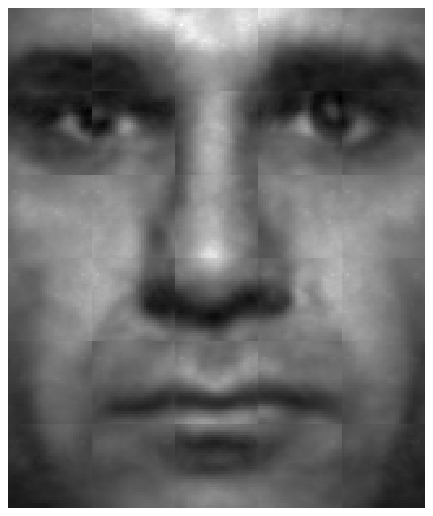}} &
 
\hspace{-.7cm}   
     \subfloat[JPEG2000 (bpp $=0.09$)]{\includegraphics[width=0.2\textwidth]{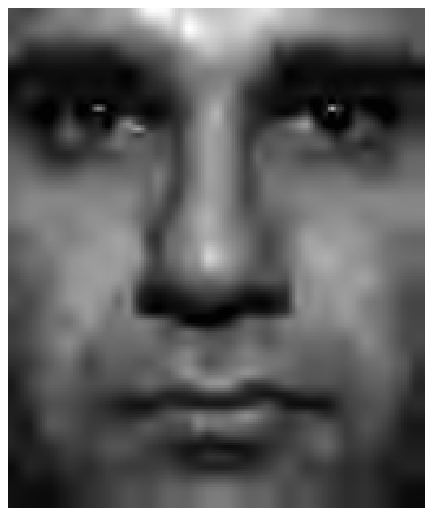}}

   \end{tabular}
   \end{center}
\caption{Visual comparison of the proposed compression with the JPEG2000 codec over random images from the test set.}
\label{fig:Visual}
\end{figure}

%
%

\section{Conclusions}
\label{sec:Summary}
We presented a multi-layer data representation approach and justified its efficiency in terms of data fidelity, memory storage and computational complexity with information-theoretic arguments. We then used this approach for the application of image compression when the images belong to a certain class, in our experiments facial images. We showed that with its simple structure in the direct pixel domain which could still be improved in different ways in terms of the choice of codebook sizes, entropy coding used or post-processing to reduce the blocking artifact,  significant performance boost was achieved in the very low rate regime, compared to the JPEG2000 codec.  
\section*{Acknowledgments}
The research has been partially supported by SNF grant 1200020-146379 and  by a grant from Switzerland through the Swiss Contribution to the enlarged European Union PSPB-125/2010.


\bibliographystyle{IEEE}
\bibliography{ArXive}

\end{document}